\documentclass[letterpaper, 10 pt, conference]{ieeeconf}  

\IEEEoverridecommandlockouts                              

\usepackage{graphicx}

\title{\LARGE \bf
Classification of Alzheimer's Disease Using fMRI Data and Deep Learning Convolutional Neural Networks
}

\author{Saman Sarraf, Ghassem Tofighi
\thanks{Data used in preparation of this article were obtained from the Alzheimer's  Disease Neuroimaging Initiative (ADNI) database (adni.loni.usc.edu). As such, the investigators within the ADNI contributed to the design and implementation of ADNI and/or provided data but did not participate in analysis or writing of this report. A complete listing of ADNI investigators can be found at: http://adni.loni.usc.edu/wp-content/uploads/how\_to\_apply/ADNI\_Acknowledgement\_List.pdf}
\thanks{Saman Sarraf was with the Department of Electrical and Computer Engineering, McMaster University, Hamilton, ON, L8S 4L8, Canada.
        {\tt\small samansarraf@ieee.org}}%
\thanks{Ghassem Tofighi is with the Electrical and Computer Engineering Department, Ryerson University, Toronto, ON M5B 2K3 Canada.
        {\tt\small gtofighi@ryerson.ca}}%
}

\begin{document}

\maketitle
\thispagestyle{empty}
\pagestyle{empty}

\begin{abstract}
Over the past decade, machine learning techniques especially predictive modeling and pattern recognition in biomedical sciences from drug delivery system \cite{sarraf2014mathematical} to medical imaging has become one of the important methods which are assisting researchers to have deeper understanding of entire issue and to solve complex medical problems. Deep learning is power learning machine learning algorithm in classification while extracting high-level features. In this paper, we used convolutional neural network to classify Alzheimer's brain from normal healthy brain.  The importance of classifying this kind of medical data is to potentially develop a predict model or system in order to recognize the type disease from normal subjects or to estimate the stage of the disease. Classification of clinical data such as Alzheimer's disease has been always challenging and most problematic part has been always selecting the most discriminative features. Using Convolutional Neural Network (CNN) and the famous architecture LeNet-5, we successfully classified functional MRI data of Alzheimer's subjects from normal controls where the accuracy of test data on trained data reached 96.85\%. This experiment suggests us the shift and scale invariant features extracted by CNN followed by deep learning classification is most powerful method to distinguish clinical data from healthy data in fMRI. This approach also enables us to expand our methodology to predict more complicated systems.
\end{abstract}

\section{INTRODUCTION}
Alzheimer's disease is a neurological, irreversible, progressive brain disorder and multifaceted disease that slowly destroys brain cells causing memory and thinking skills loss, and ultimately the ability to carry out the simplest tasks. The cognitive decline caused by this disorder ultimately leads to dementia. For instance, the disease begins with mild deterioration and gets progressively worse in a neurodegenerative type of dementia. Diagnosing Alzheimer's disease requires very careful medical assessments such as patients? history, Mini Mental State Examination (MMSE) and physical and neurobiological exam. In addition to those evaluations, resting state functional magnetic resonance imaging (rs-fMRI) provides a non-invasive method to measure functional brain activity and changes in the brain \cite{vemuri2012resting}. There are two important concepts about resting-state fMRI. First of all, as patients do not need to do any task and there is no simulation, the procedure will be more comfortable than a normal fMRI and secondly, rs-fMRI data acquisition can be performed during a clinical scan and most researchers are interested in brain network analysis and extraction from rs-fMRI data  \cite{sarraf2016functional} \cite{sarraf2016robust} \cite{strother2014hierarchy} \cite{sarraf2014brain} \cite{grady2016age}. However, development of an assistant tool or algorithm to classify fMRI data and more importantly to recognize brain disorder data from healthy subjects has been always clinicians? interests. Any machine learning algorithm which is able to classify Alzheimer's disease assists scientists and clinicians to diagnose this brain disorder. In this work, the convolutional neural network (CNN) which is one of the Deep Learning Network architecture is utilized in order to classify the Alzheimer's brains and healthy brains and to produce a trained and predictive model.

\section{Background and Algorithms}
\subsection{Data Acquisition and Preprocessing}

For this study, Alzheimer`s Disease (AD) patients and 15 elderly normal control subjects (24 female and 19 male) with a mean age of 74.9 ± 5.7 years were selected from ADNI dataset. The AD patients had MMSE over 20 reported by *ADNI and all normal participants were healthy and had no reported history of medical or neurological conditions. Scanning was performed on a Siemens Trio 3 Tesla MRI scanner.  Anatomical scans were acquired with a 3D MP-RAGE sequence (TR=2s, TE=2.63 ms, FOV=25.6 cm, 256 x 256 matrix, 160 slices of 1mm thickness). Functional runs were obtained with an EPI sequence (150 volumes, TR=2 s, TE=30 ms, flip angle=70?, FOV=20 cm, 64 x 64 matrix, 30 axial slices of 5mm thickness, no gap).
The fMRI data were pre-processed using the standard modules of FMRIB Software Library v5.0 \cite{smith2004advances}. The pre-processing steps for the anatomical data involved the removal of non-brain tissue from T1 anatomical images using the Brain Extraction Tool. For functional data the preprocessing steps included: motion correction (MCFLRIT), skull stripping, and spatial smoothing (Gaussian kernel of 5-mm FWHM). Low-level noise was removed using a high-pass temporal filtering (?=90.0 sec). The functional images were then aligned to the individual?s high-resolution T1-weighted scans, which were subsequently registered to the Montreal Neurological Institute standard space (MNI152) using affine linear registration and resampled at 2mm cubic voxels. The product of preprocessing step was 45x54x45x300 images in which the first 10 slices of each image were removed as they contained no functional information. 

\subsection{Deep Learning}

Hierarchical or structured deep learning is a modern branch of machine learning that was inspired by human brain. It has been developed based on complicated algorithms that model high-level features and extract those abstractions from data by using similar neural network architecture but much complicated. The neuroscientists discovered the ?neocortex? which is a part of the cerebral cortex concerned with sight and hearing in mammals, process sensory signals by propagating them through a complex hierarchy over time.  That was the main motivation to develop the deep machine learning focusing on computational models for information representation that exhibit similar characteristics to that of the neocortex \cite{arel2010deep} \cite{lecun1998gradient} \cite{jia2014caffe}.
\subsubsection{Convolutional Neural Networks (CNNs / ConvNets)}

Convolutional Neural Networks which are inspired by human visual system are similar to classic neural networks. This architecture has been particularly designed based on the explicit assumption that raw data are two-dimensional (images) that enables us to encode certain properties and also to reduce the amount of hyper parameters. The CNN topology utilizes spatial relationships to reduce the number of parameters which must be learned and thus improves upon general feed-forward back propagation training. Equation \ref{error_eqn} shows how Error is calculated in the back propagation step where E is error function, y is the ith, jth neuron,  x is the input, l represent layer numbers w is filter weight with a and b indices, N is the number of neurons in a given layer and m is the filter size.

\begin{equation}\label{error_eqn1}
 \frac{\partial E}{\partial  W_{ab} } =  \sum_{i=0}^{N-m}\sum_{j=0}^{N-m}\frac{\partial E}{\partial  W_{ij}^l }\frac{\partial x_{ij}^l}{\partial  W_{ab}}
\end{equation}
\begin{equation}\label{error_eqn2}
 =\sum_{i=0}^{N-m}\sum_{j=0}^{N-m}\frac{\partial E}{\partial  x_{ij}^l }  y_{(i+a)(j+b)} ^{l-1} 
\end{equation}
In CNNs, small portions of the image (dubbed a local receptive field) are treated as inputs to the lowest layer of the hierarchical structure. One of the most important features of CNN is that complex architecture provides a level of invariance to shift, scale and rotation as the local receptive field allows the neuron or processing unit access to elementary features such as oriented edges or corners. 
This network is basically made up of neurons having learnable weights and biases forming Convolutional Layer. It also includes other network structures such as Pooling Layer, Normalization Layer and Fully-Connected Layer. As briefly mentioned above, Convolutional or so called CONV layer computed the output of neurons that are connected to local regions in the input, each computing a dot product between their weights and the region they are connected to in the input volume. Pooling or so called POOL Layer performs a down sampling operation along the spatial dimensions. The Normalization Layer or RELU layer applies an elementwise activation function, such as the max (0, x) thresholding at zero. This layer does not change the size of the image volume. Fully-Connected Layer (FC) layer computes the class scores, resulting in volume of number of classes. As with ordinary Neural Networks and as the name implies, each neuron in this layer will be connected to all the numbers in the previous volume \cite{arel2010deep} \cite{lecun1998gradient}.
The Convolutional Layer plays an important role in CNN architecture and is the core building block in this network. The CONV layer's parameters consist of a set of learnable filters. Every filter is spatially small but extends through the full depth of the input volume. During the forward pass, each filter is convolved across the width and height of the input volume, producing a 2D activation map of that filter. During this convolving, the network learns filters that activate when they see some specific type of feature at some spatial position in the input. Next, these activation maps are stacked for all filters along the depth dimension forms the full output volume. 
Every entry in the output volume can thus also be interpreted as an output of a neuron that looks at only a small region in the input and shares parameters with neurons in the same activation map \cite{arel2010deep} \cite{lecun1998gradient} \cite{jia2014caffe}.
A Pooling layer is usually inserted in-between successive Conv layers in ConvNet architecture. Its function is to reduce (down sample) the spatial size of the representation in order to reduce the amount of network hyper parameters, and hence to also control overfitting. The Pooling Layer operates independently on every depth slice of the input and resizes it spatially, using the MAX operation. Recently, more successful CNN have been developed such as LeNet, AlexNet, ZF Net, GoogleNet, VGGNet and ResNet. The major bottleneck of constructing ConvNet architectures is the memory restrictions of GPU \cite{arel2010deep} \cite{lecun1998gradient} \cite{jia2014caffe}.
\subsubsection{Adoped LeNet-5}
As you can see in the figure \ref{fig:lenet}, LeNet-5  was firstly designed by Y. LeCun et Al. \cite{lecun1998gradient} and this famous network successfully classified digits and was applied to hand-written check numbers. The application of this network was expanded to more complicated problems and their hyper parameters adjusted for new issues. However, more developed versions of LeNet have been successfully tested. In this work, we dealt with a binary but very complicated binary classification of Alzheimer's and Normal data. In other word, we needed a complicated network but for two classes which was leading us to choose LeNet-5 and adjust this architecture for fMRI data.
\begin{figure}[h]
	\centering
	\includegraphics[width=.50\textwidth,natwidth=1824,natheight=452]{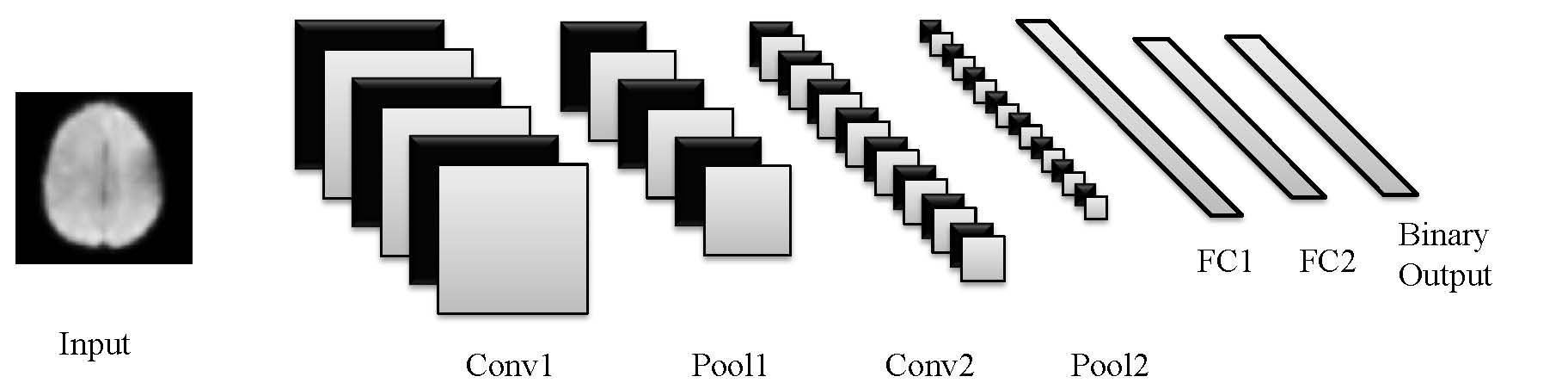}
	\caption{LeNet-5 architecture adopted for fMRI data}
	\label{fig:lenet}
\end{figure}

\begin{figure}[h]
	\centering
	\includegraphics[width=.20\textwidth,natwidth=448,natheight=2157]{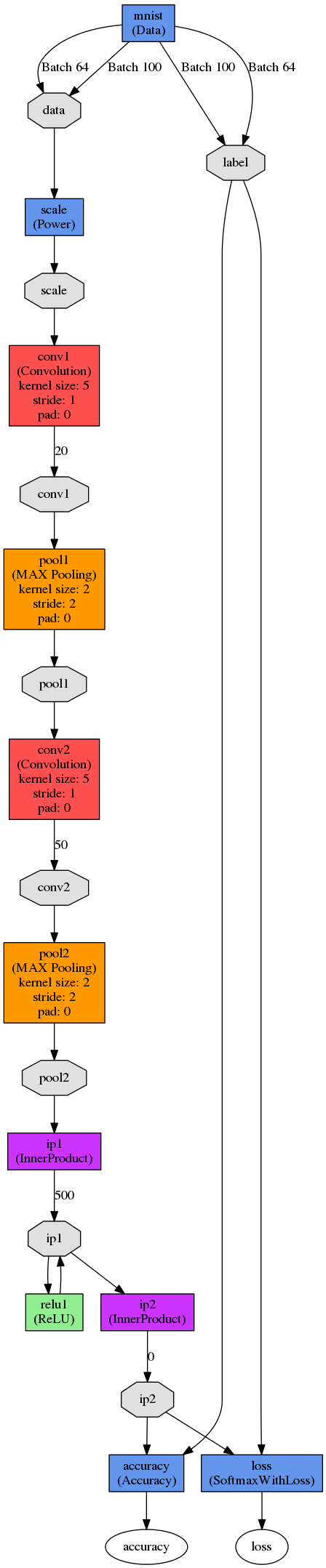}
	\caption{LeNet-5 network implemented for fMRI data}
	\label{fig:network}
\end{figure}

The implemented network is shown in figure \ref{fig:network} in details.

\section{Results and Discussion}
The preprocessed fMRI 4D data in Nifti format were concatenated across z and t axes and the converted to a stack of 2D images in JPEG using Neuroimaging package Nibabel (http://nipy.org/nibabel/) and Python OpenCV (opencv.org). Next, images were labeled for binary classification of Alzheimer's Vs Normal data. The labeled images were converted to lmdb storage Databases for high-throughput to be fed into Deep Learning platform. LeNet model which is based on Convolutional Neural Network architecture from Caffe DIGITS 0.2 - deep learning framework (Nvidia version) - was used to perform binary image classification. The data was divided into three parts: training (60\%), validation (20\%) and testing (20\%). The number of epochs was set to 30 and the batch size was 64 resulting in 126990 iterations. The LeNet was trained by 270900 samples and validated and tested by 90300 images. In order to achieve the robustness and reproducibility of the deep neural network, the cross validation process was repeated 5 times (5-fold cross validation) and gathered in table \ref{table:1}.
\begin{table}[h!]
	\centering
	\caption{Accuracies achieved from CNN across 5 runs}
	\begin{tabular}{|l|l|l|l|l|c|}
		\hline
		
		Run1    & Run2 & Run3 & Run4 & Run5 & \textbf{Mean} \\
		\hline
		96.858 & 96.857  & 96.854 & 96.863 & 96.8588 & \textbf{96.8588}\\
		\hline
	\end{tabular}
	\label{table:1}
\end{table}
The mean of images removed from data in the Deep Learning preprocessing step. In the training phase, loss of training and testing and also accuracy of validation data were measured. Learning rate dramatically fell down after the 10th epoch and slightly decreased after 20th epoch as is shown in. Deep Learning LeNet model successfully recognized the Alzheimer's data from Normal Control and the averaged accuracy reached 96.8588\% as it is depicted in figure \ref{fig:lenet}.  
The training and testing process were performed on NVIDIA GPU Cloud Computing which significantly improved the performance of Deep Learning classifier.

\begin{figure}[h]
	\centering
	\includegraphics[width=.50\textwidth,natwidth=1564,natheight=1115]{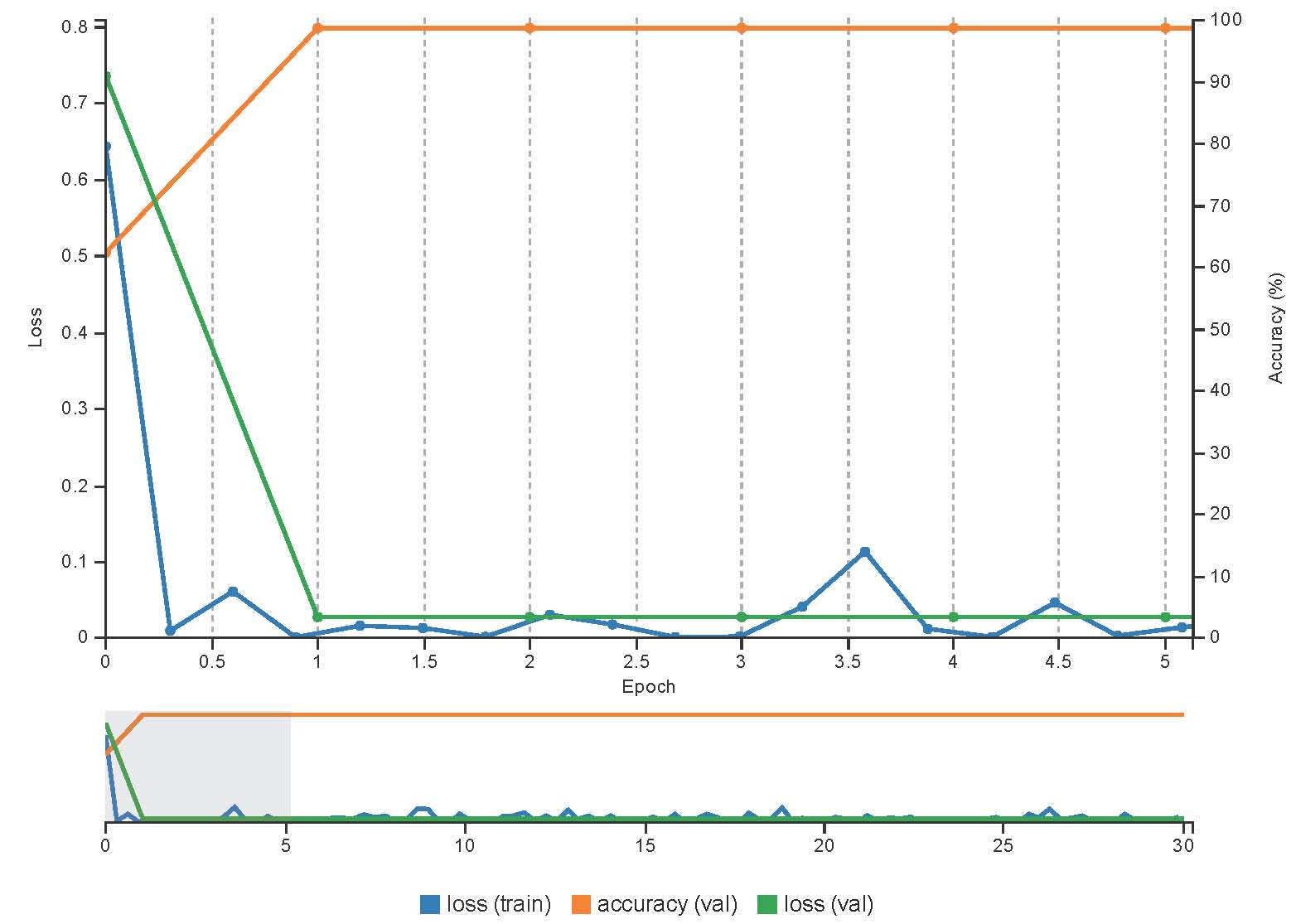}
	\caption{Loss train, Accuracy Validation (Test), Loss Validation for Run5}
	\label{fig:lenet}
\end{figure}

\begin{figure}[h]
	\centering
	\includegraphics[width=.40\textwidth,natwidth=1583,natheight=692]{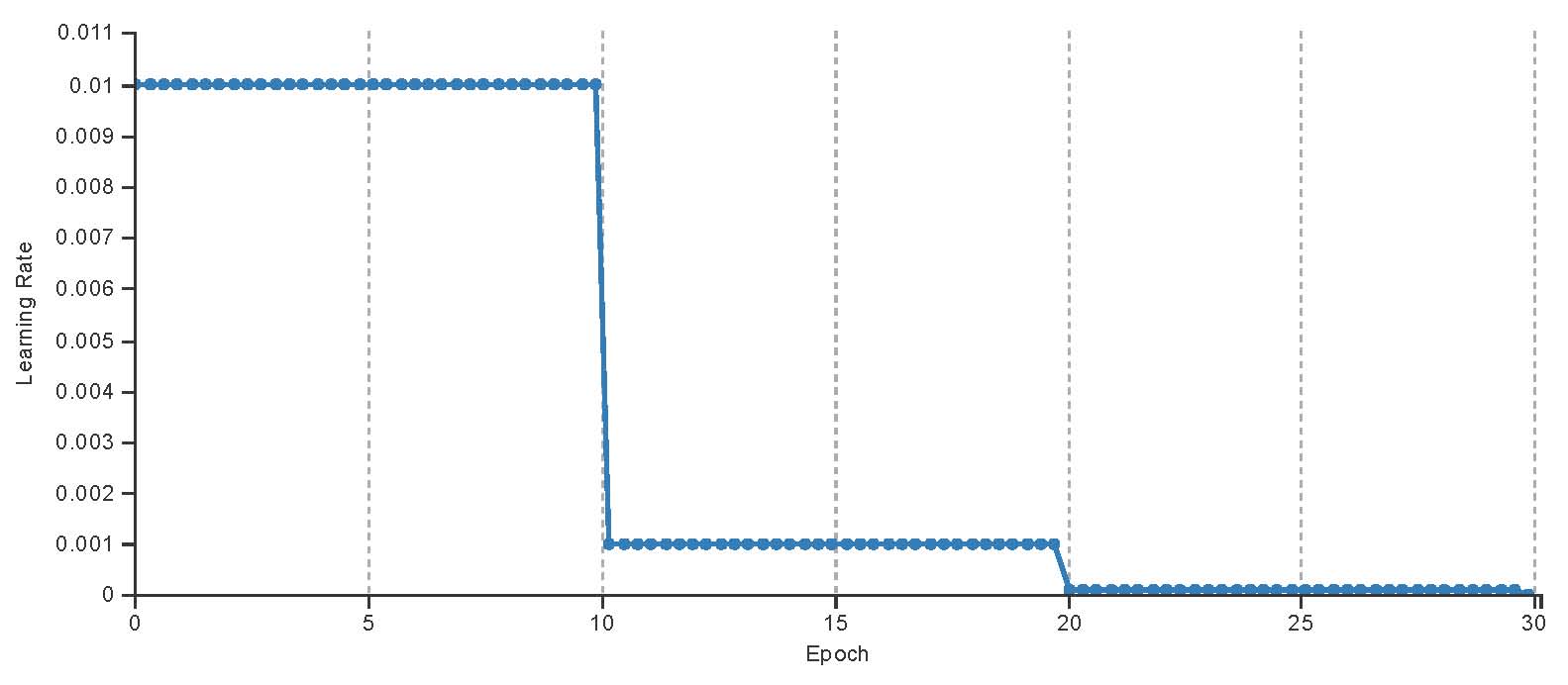}
	\caption{Learning Rate dropping each 10 epochs}
	\label{fig:learning-rate}
\end{figure}

Figure \ref{fig:learning-rate} shows how learning rate is dropping each 10 epochs in our training sets. We have started with 0.01 and divided it by 10 in each 10 epochs.

Most challenges in traditional medical image processing and analyses have been to select the best and most discriminative features that must be extracted from data and choosing the best classification method. Two important advantages of Deep Learning methods especially Convolutional Neural Network used in this work is to contain those two characteristics simultaneously. It is possible to visualize the results of filters (kernels) in each layer. 

\begin{figure}[h]
	\centering
	\includegraphics[width=.38\textwidth,natwidth=834,natheight=981]{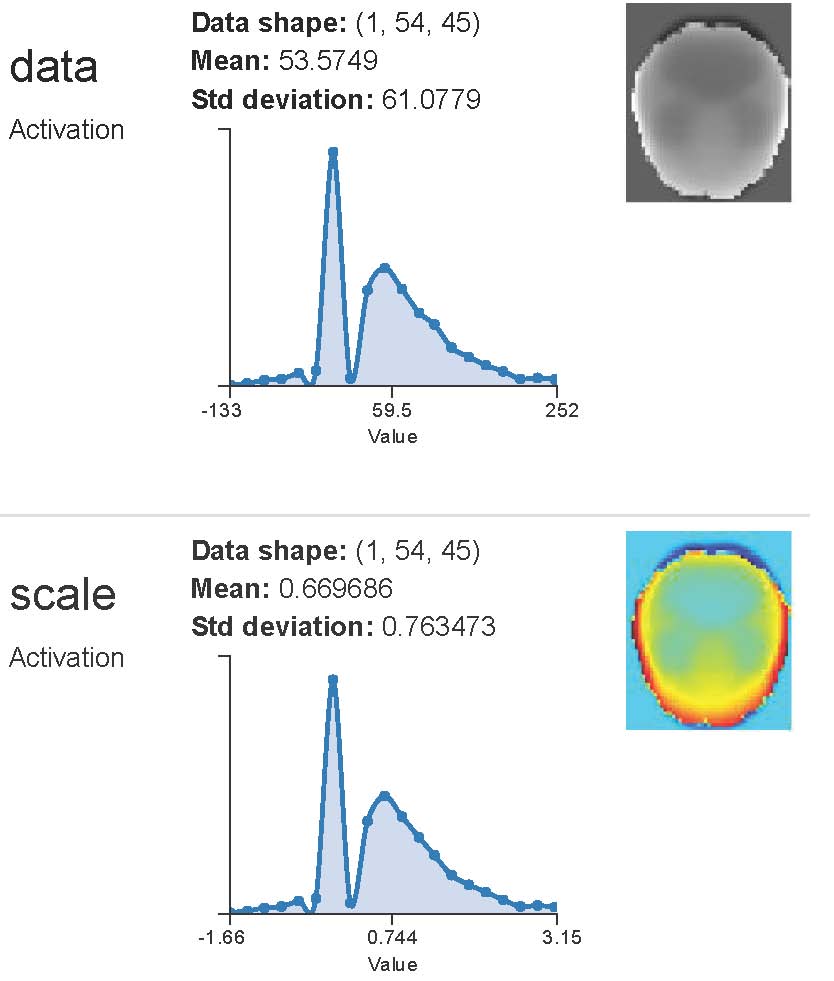}
	\caption{Statistics and Visualization for data and scale of Alzheimer's Disease sample}
	\label{fig:ad-data}
\end{figure}

\begin{figure}[h]
	\centering
	\includegraphics[width=.38\textwidth,natwidth=862,natheight=980]{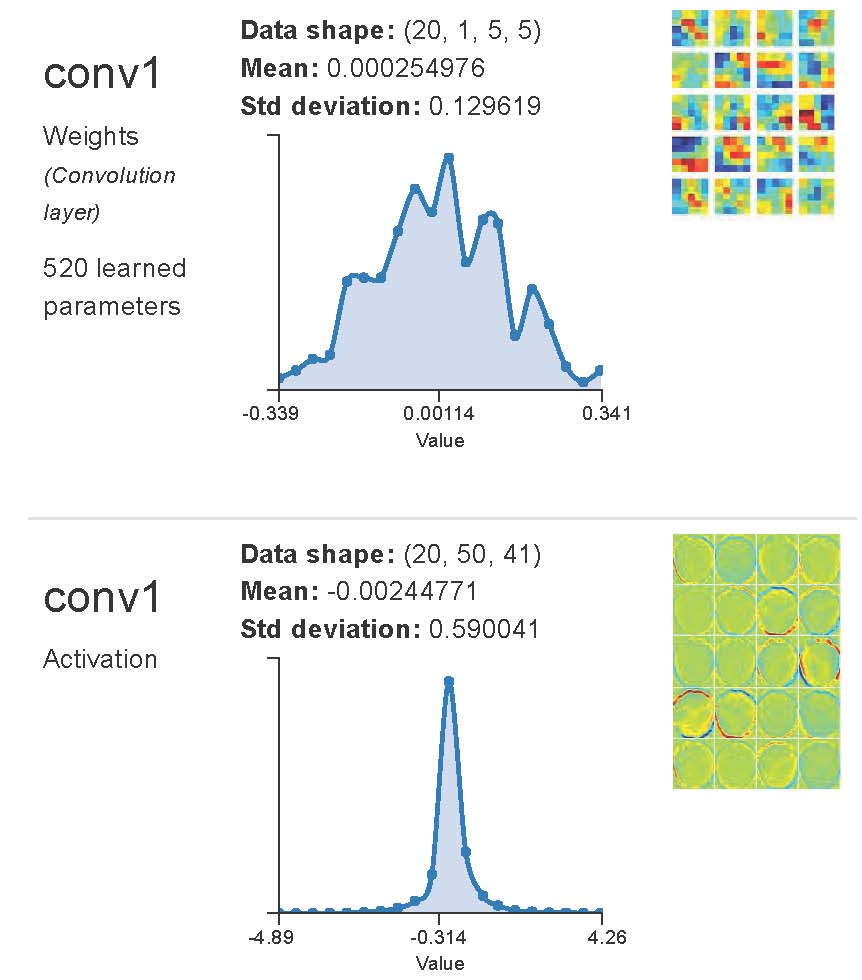}
	\caption{Statistics and visualization for first convolution layer of Alzheimer's Disease sample}
	\label{fig:ad-conv1}
\end{figure}

\begin{figure}[h]
	\centering
	\includegraphics[width=.38\textwidth,natwidth=983,natheight=1118]{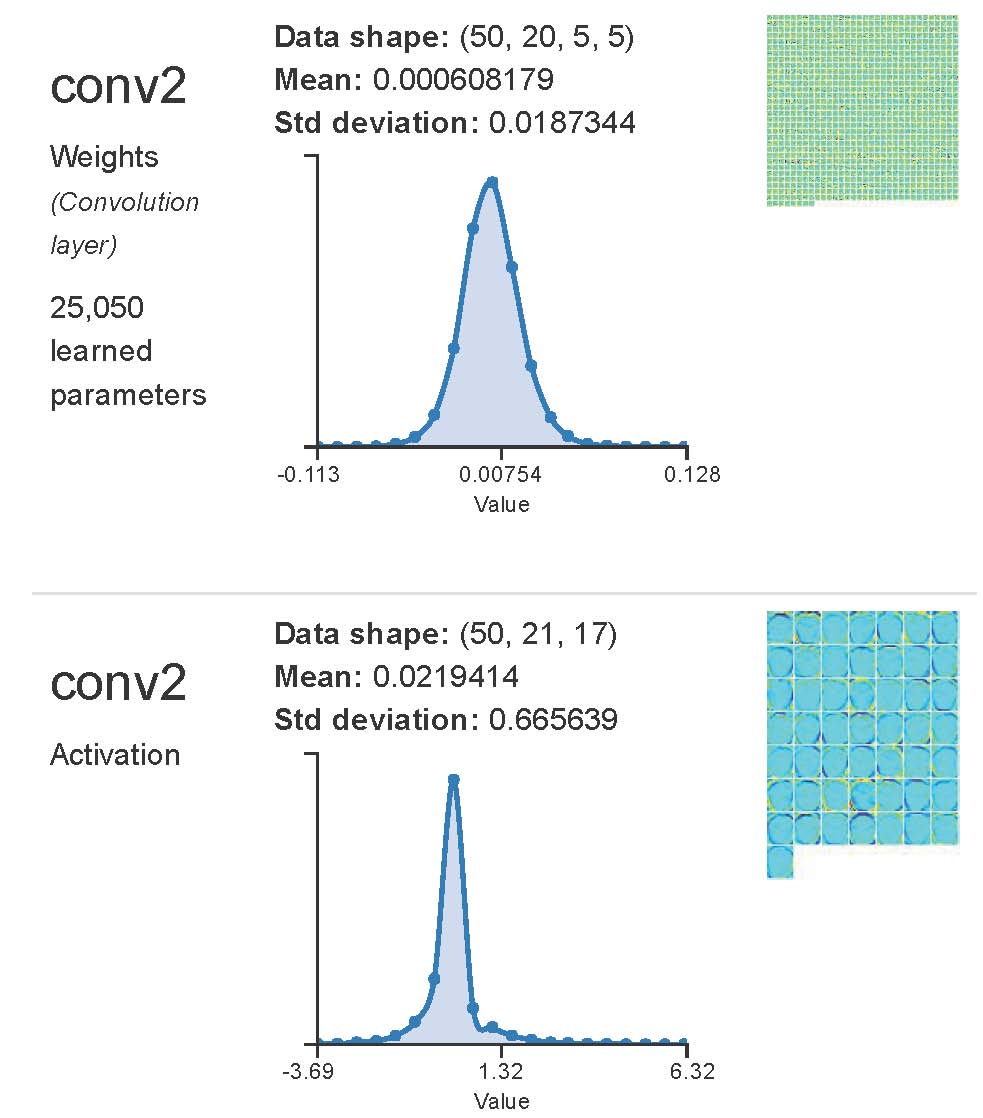}
	\caption{Statistics and visualization  for second convolution layer of Alzheimer's Disease sample}
	\label{fig:ad-conv2}
\end{figure}

Figures \ref{fig:ad-data}, \ref{fig:ad-conv1} and \ref{fig:ad-conv2} illustrate some of the final filter of different layers. CNN is a strong feature extractor because of its convolutional layers that is able to extract high level features from images. This deep learning method is also a powerful classifier because of its complicated network architecture. Our current solution based on fully advanced preprocessing steps followed by CNN classification improved the accuracy of AD data classification from 84\% using Support Vector Machine (SVM) reported in the literature \cite{zhang2015resting} \cite{raventosautomating} \cite{tripoliti2008supervised} to 96.86\%. However, the deep learning solutions have very few problems such as high algorithm complexity and expensive infrastructure.

\section{Conclusions}
In this paper, we successfully classified the AD data from normal control with 96.86\% accuracy using CNN deep learning architecture (LeNet) which was trained and tested with huge number of images. This deep learning solution and our proposed pipeline not only open new avenues in medical image analyses but also it enables researchers and physicians to potentially predict any new data. It is also possible to generalize this method to predict different stages of Alzheimer's disease for different age groups. Furthermore, this deep learning-based solution allows researchers to perform feature selection and classification with a unique architecture. The accuracy achieved in this work was very high confirming the network architecture was correctly selected. However, more complicated network architecture including more convolutional neural layers is recommended for future works and more complicated problems.
\section{Acknowledgement}
We would like to express our gratitude towards Dr. Cristina Saverino, post-doctoral fellowship at Toronto Rehabilitation Institute-University Health Network and Department of Psychology - University of Toronto and for extending her help and support in this study.

\section{Appendix I}
Data collection and sharing for this project was funded by the Alzheimer's Disease Neuroimaging Initiative (ADNI) (National Institutes of Health Grant U01 AG024904) and DOD ADNI (Department of Defense award number W81XWH-12-2-0012). ADNI is funded by the National Institute on Aging, the National Institute of Biomedical Imaging and Bioengineering, and through generous contributions from the following: AbbVie, Alzheimer?s Association; Alzheimer?s Drug Discovery Foundation; Araclon Biotech; BioClinica, Inc.; Biogen; Bristol-Myers Squibb Company; CereSpir, Inc.; Eisai Inc.; Elan Pharmaceuticals, Inc.; Eli Lilly and Company; EuroImmun; F. Hoffmann-La Roche Ltd and its affiliated company Genentech, Inc.; Fujirebio; GE Healthcare; IXICO Ltd.; Janssen Alzheimer Immunotherapy Research \& Development, LLC.; Johnson \& Johnson Pharmaceutical Research \& Development LLC.; Lumosity; Lundbeck; Merck \& Co., Inc.; Meso Scale Diagnostics, LLC.; NeuroRx Research; Neurotrack Technologies; Novartis Pharmaceuticals Corporation; Pfizer Inc.; Piramal Imaging; Servier; Takeda Pharmaceutical Company; and Transition Therapeutics. The Canadian Institutes of Health Research is providing funds to support ADNI clinical sites in Canada. Private sector contributions are facilitated by the Foundation for the National Institutes of Health (www.fnih.org). The grantee organization is the Northern California Institute for Research and Education, and the study is coordinated by the Alzheimer's Disease Cooperative Study at the University of California, San Diego. ADNI data are disseminated by the Laboratory for Neuro Imaging at the University of Southern California.

\end{document}